\documentclass[letterpaper, 10 pt, conference]{ieeeconf}  

\usepackage{multicol}
\usepackage{lmodern}
\usepackage{xcolor}
\usepackage{newunicodechar}
\newunicodechar{，}{,}
\newunicodechar{（}{(}
\newunicodechar{）}{)}
\definecolor{DodgerBlue}{RGB}{30,144,255}
\usepackage{caption}
\usepackage{makecell}

\usepackage{graphics} 
\usepackage{epsfig} 
\usepackage{times} 
\usepackage{amsmath} 
\usepackage{amssymb}  
\usepackage[utf8]{inputenc}
\usepackage[ruled,vlined, linesnumbered]{algorithm2e}
\usepackage[margin=1in]{geometry}
\usepackage{subcaption}
\usepackage{booktabs}
\usepackage{multirow}
\usepackage{array}
\usepackage[misc]{ifsym}
\usepackage{array} 
\newcolumntype{L}[1]{>{\raggedright\arraybackslash}p{#1}}

\usepackage{amsmath} 
\usepackage{xparse}
\usepackage{mathtools} 
\usepackage{booktabs} 
\ExplSyntaxOn
\NewDocumentCommand{\definealphabet}{mmmm}
 {
  \int_step_inline:nnn { `#3 } { `#4 }
   {
    \cs_new_protected:cpx { #1 \char_generate:nn { ##1 }{ 11 } }
     {
      \exp_not:N #2 { \char_generate:nn { ##1 } { 11 } }
     }
   }
 }
\ExplSyntaxOff
\definealphabet{bb}{\mathbb}{A}{Z}
\definealphabet{bf}{\mathbf}{A}{Z}
\definealphabet{mc}{\mathcal}{A}{Z}
\definealphabet{mf}{\mathfrak}{A}{Z}
\definealphabet{mf}{\mathfrak}{a}{z}

\usepackage{amsfonts}

\makeatletter
\let\NAT@parse\undefined
\makeatother
\usepackage[colorlinks=true]{hyperref}
\hypersetup{breaklinks=true,bookmarks=true,colorlinks=true,linkcolor=DodgerBlue,urlcolor=DodgerBlue, citecolor=DodgerBlue,}
\title{\LARGE \bf 
Generalizable Hierarchical Skill Learning via Object-Centric Representation
}

\author{
Haibo Zhao\textsuperscript{1}, 
Yu Qi\textsuperscript{1},
Boce Hu\textsuperscript{1},
Yizhe Zhu\textsuperscript{1},
Ziyan Chen\textsuperscript{1},
Heng Tian\textsuperscript{1},
Xupeng Zhu\textsuperscript{1},\\
Owen Howell\textsuperscript{1},
Haojie Huang\textsuperscript{1},
Robin Walters\textsuperscript{1}, 
Dian Wang\textsuperscript{2\dag}, 
Robert Platt\textsuperscript{1\dag}\\
\textsuperscript{1}Northeastern University \quad\quad
\textsuperscript{2}Stanford University \quad\quad \textsuperscript{\dag}Equal Advising \\
\texttt{\url{ https://codemasterzhao.github.io/GSL/}}
}

\begin{document}
\maketitle
\begin{abstract}
We present Generalizable Hierarchical Skill Learning (GSL), a novel framework for hierarchical policy learning that significantly improves policy generalization and sample efficiency in robot manipulation. 
One core idea of GSL is to use object-centric skills as an interface that bridges the high-level vision-language model and the low-level visual-motor policy.
Specifically, GSL decomposes demonstrations into transferable and object-canonicalized skill primitives using foundation models, ensuring efficient low-level skill learning in the object frame.
At test time, the skill–object pairs predicted by the high-level agent are fed to the low-level module, where the inferred canonical actions are mapped back to the world frame for execution.
This structured yet flexible design leads to substantial improvements in sample efficiency and generalization of our method across unseen spatial arrangements, object appearances, and task compositions. In simulation, GSL trained with only 3 demonstrations per task outperforms baselines trained with 30 times more data by 15.5\% on unseen tasks. In real-world experiments, GSL also surpasses the baseline trained with 10 times more data.
\end{abstract}

\section{Introduction}

Recent advances in Vision-Language Models (VLMs) have enabled intelligent agents to interpret rich semantic instructions and perform diverse visual and language tasks~\cite{qi2025bear}. Building upon these models, a growing body of work seeks to enhance the generalization ability of robotic manipulation policies by leveraging the semantic grounding and compositional reasoning inherent to VLMs. Among these efforts, a prevalent line of research has focused on developing Vision-Language-Action (VLA) models \cite{kim2024openvla, intelligence2504pi0, zhou2025vision}, which couple perception, language, and control into a unified framework trained on large-scale robot datasets. By mapping visual-language inputs directly to low-level actions, these approaches demonstrate promising semantic and visual understanding ability.
However, despite their scalability, the performance of VLA models remains limited by the sparsity of robotic data, as well as the difficulty of learning robust and generalizable visuomotor mappings end-to-end from high-dimensional inputs.
\begin{figure}[ht!]
    \centering
    \includegraphics[width=0.8\columnwidth]{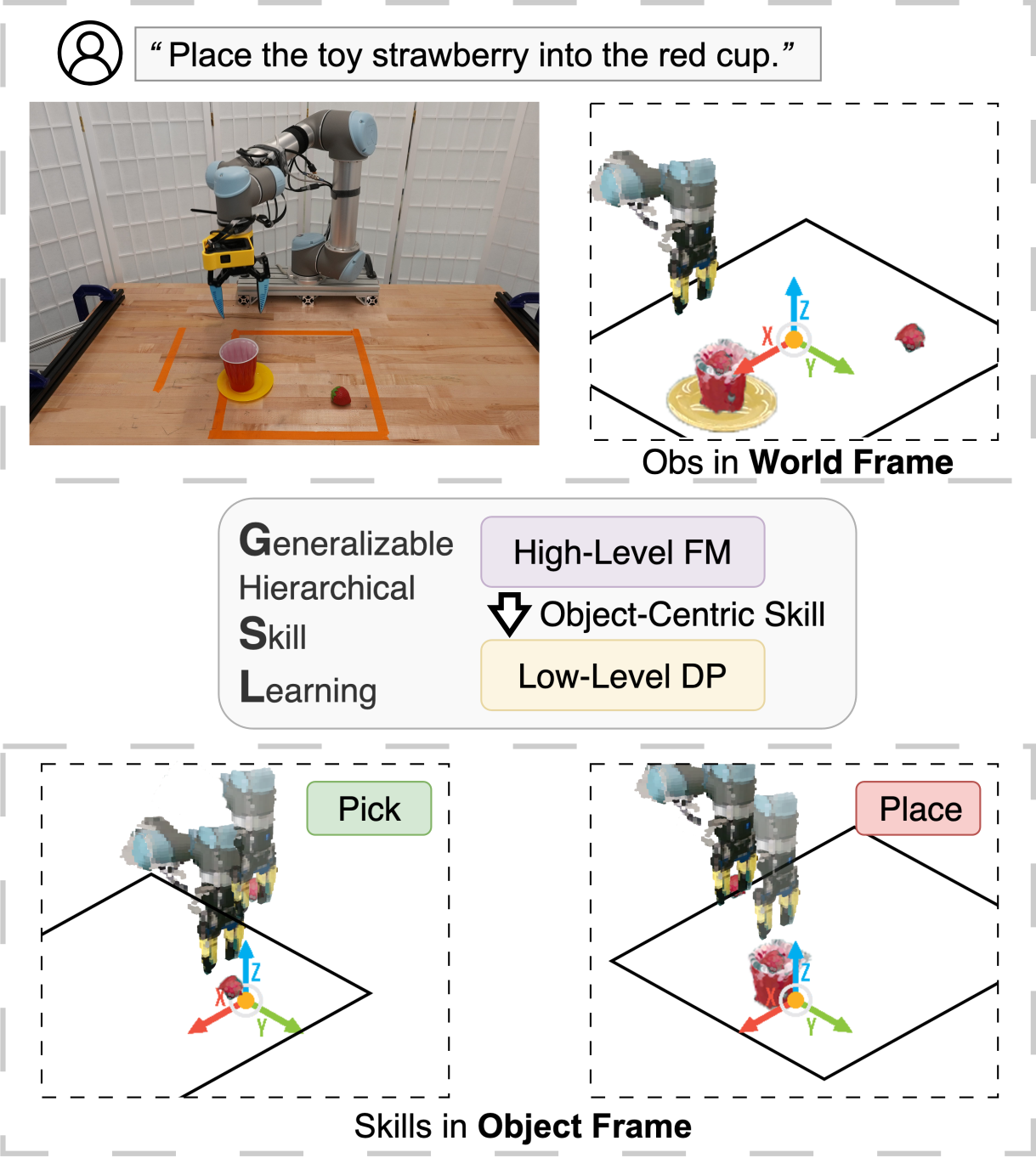}
    \caption{\textbf{Overview of GSL.} In GSL, the high-level agent decomposes demonstrations into object-centric skills for the low-level module to learn. At test time, the high-level agent selects a sequence of skills and object pairs to execute novel tasks. }
    \label{fig:fig1}
    \vspace{-0.5cm}
\end{figure}
Alternative strategies introduce a hierarchical structure into policy learning~\cite{levy2018hierarchical, ma2024hierarchical, zhao2025hierarchical,garcia2025towards}. Instead of directly predicting low-level actions, hierarchical approaches decompose the control process into high-level task planning and low-level execution. The high-level planner, often realized by a VLM, predicts abstract subgoals or symbolic commands, while the low-level controller executes precise visuomotor behaviors conditioned on those subgoals. Compared with monolithic VLA models, this hierarchical formulation embeds a stronger inductive bias: it allocates semantic reasoning to the VLM, which excels at task abstraction, and constrains the visuomotor policy to focus on local control, thereby reducing learning complexity and improving robustness. Nevertheless, existing hierarchical policies typically employ simple and coarse interfaces between the high-level and low-level modules, such as object masks, keyframes, or end-effector poses. While such representations facilitate modularity, they provide limited structural information for the low-level controller, making it challenging to learn fine-grained skills and to generalize across task and object configurations.


To bridge this gap, we propose a novel hierarchical framework that introduces object-centric skills as the interface between high-level reasoning and low-level control. During training, demonstrations are decomposed into object-centric skills expressed in the object frame, forming an object-centric dataset that enables the low-level policy to learn transferable visuomotor primitives grounded in consistent spatial references. At inference time, the high-level agent decomposes a long-horizon task into a sequence of object-centric skills by predicting the skill types and the corresponding objects, while the low-level policy generates the corresponding skill trajectory within the object frame. This formulation establishes a structured yet flexible communication channel, allowing the high level to convey semantic and spatial intent, and the low-level module to perform localized control without reasoning over the entire scene. In addition, our method achieves substantially higher sample efficiency and stronger generalization to novel objects and tasks compared to flat or weakly structured baselines across tasks involving spatial variation, novel object appearance, distractor objects, and unseen task combinations, regarded as key aspects of generalization in~\cite{garcia2025towards,jiang2023vima}.


Our contributions are summarized as follows:  
\begin{itemize}
\item We propose \textbf{G}eneralizable Hierarchical \textbf{S}kill \textbf{L}earning (GSL), a novel framework for generalizable hierarchical policy learning.
\item We introduce object-centric skill as an interface between high-level and low-level, as a structured yet flexible communication channel.
\item  We empirically show that GSL significantly improves generalization over strong baselines, with GSL trained on only 3 demonstrations outperforming baselines trained on 100 demonstrations by 15.5\% on unseen tasks.
\end{itemize}

\section{Related Work}

\textbf{Hierarchical Policy Learning.} Hierarchical policy strategy decomposes a policy into high-level and low-level agents for both long-horizon reasoning and precise fine-grained control~\cite{levy2018hierarchical, gualtieri2020learning, sharma2017learning, zhao2025hierarchical, zhu2022sample, ma2024hierarchical, hu2025push}. Recent works leverage vision-language models at a high level for improving generalization abilities in imitation learning~\cite{fang2025sam2act,ke20243d,gervet2023act3d,zhou2025vision,intelligence2504pi0, shridhar2022cliport, kim2024openvla}. We highlight two series of works among them, the first passes object grounding to low-level policy for fine-grained actions, like BridgeVLA~\cite{li2025bridgevla} uses VLM as the backbone of high-level agents to predict heatmaps on 2D images of 3D projections. The second decomposes tasks into skills and recombines them into novel tasks, like 3D-Lotus~\cite{garcia2025towards}. 
However, these approaches typically lack explicit object-centric interfaces, 
making their skill recomposition less robust to environmental and geometric variations.
 Different from the above methods, we leverage an object mask as a reference frame and use object-centric skills composition for generalizable policy learning.

\textbf{Object-centric Policy Learning.} Object-centric formulations improve the generalization ability of imitation learning by grounding perception and action in task-relevant objects. Many manipulation skills depend on object geometry and affordances while remaining invariant to scene changes. Object-centric pipelines ground goals to objects via detection~\cite{ chapin2025object, li2025language}, part parsing~\cite{qi2025two, geng2023partmanip, li2024unidoormanip} or keypoint estimation~\cite{sundaresan2023kite,tang2025functo, huang2024rekep, goyal2023rvt,fang2024keypoint,qian2024task,li2025bridgevla}, sometimes using slot encoders~\cite{zhu2023learning} or scene graphs~\cite{jiang2024roboexp} to represent entities and relations. In our work, we define object-centric skills, as a tuple of a skill name, a canonical object point cloud expressed in the object frame, and a canonical trajectory, enabling generalization in long-horizon policy learning. 

\textbf{Equivariant Policy Learning.} Recent research has leveraged geometric symmetries in 3D Euclidean space to enhance manipulation policies~\cite{ kim2023se, yang2024equibot, zhu2025equact, hu2024orbitgrasp, hu20253d,wang2022robot,wang2020policy}. A broad range of studies explore this principle across settings such as bi-equivariant pick-and-place control~\cite{simeonov2022neural,huang2024fourier, huang2024imagination}, equivariant grasp learning~\cite{zhu2022sample, huang2022edge}, and trajectory learning~\cite{wang2024equivariant,zhu2025se} for fine-grained manipulation. However, previous works have primarily focused on scene-level symmetry, where symmetry is preserved only under transformations applied to the entire scene. In contrast, our approach emphasizes object-level symmetry by leveraging the object mask as an interface and an equivariant network for equivariant skill learning. This enables robust performance across variations in object geometry and placement, even when the background does not transform accordingly.

\section{Problem Definition}
In this work, we study hierarchical visuomotor policy learning for robotic manipulation. 
Formally, we aim to learn a policy $\pi: O \to A$ that maps from an observation space $O$ to an action space $A$.

The observation space $O$ consists of a predefined skill set $\mathcal{K}$, a task description $d$, and multi-view RGBD images $I = \{ I_v \in \mathbb{R}^{4 \times H \times W} \mid v=1,\dots,V \}$, where each $I_v$ is a four-channel image (RGB + depth).

The robot action is represented as $s = (x,y,z,q,c) \in S = \mathbb{R}^3 \times SO(3) \times \mathbb{R}$, where $(x,y,z)$ denotes the gripper position, $q$ the orientation, and $c$ the gripper aperture. The action space $A$ is defined as a trajectory $\tau = \{ s_1, s_2, \dots, s_m \} \in S^m = A$, representing $m$ control steps of the gripper state.

\section{Method}
\begin{figure*}[t]
    \vspace{0cm}
    \begin{center}
    \includegraphics[width=1\textwidth]{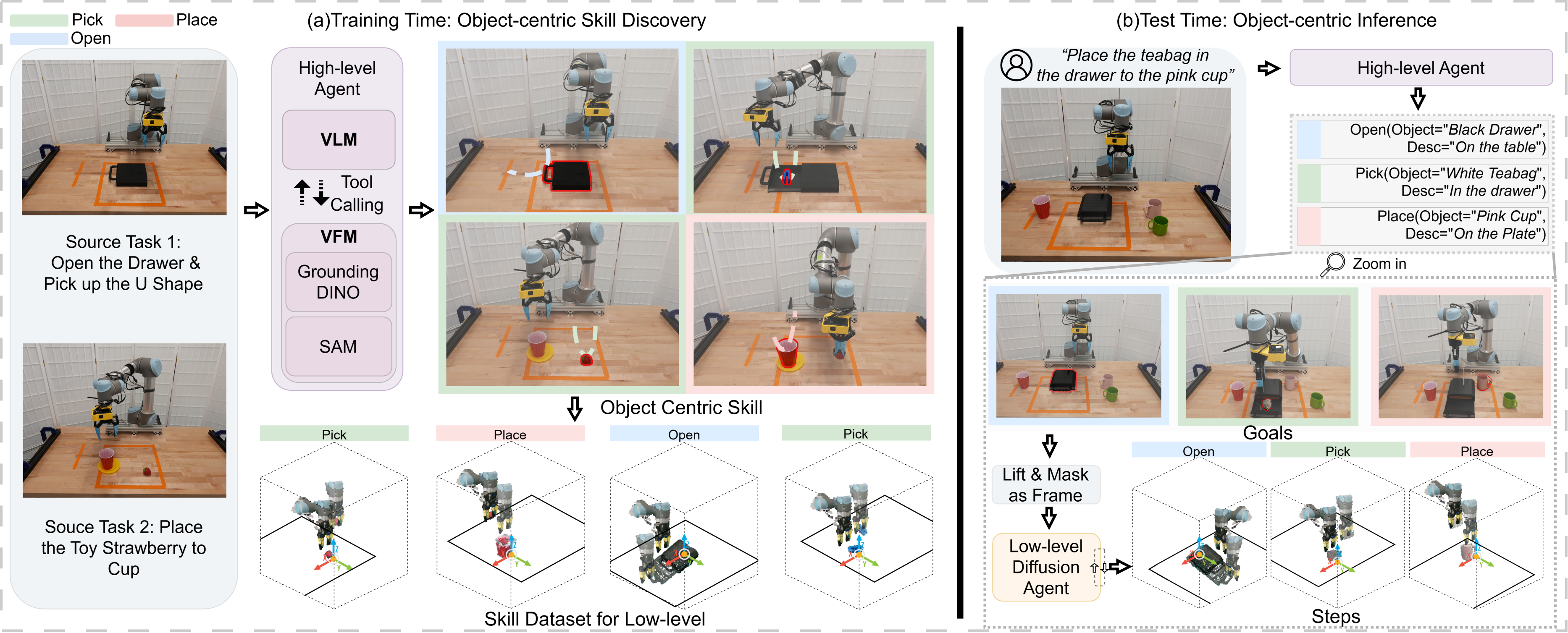}
    \vspace{-0cm}
    \caption{\textbf{Illustration of the GSL Workflow.} In the training process (left), given a demonstration consisting of a task description, multi-view RGBD frames, and ground-truth trajectories, the high-level agent decomposes the demo into object-centric skills, each defined by an object mask, a skill label, and a trajectory segment. The mask is lifted into an object point cloud, and both the point cloud and trajectory are canonicalized into the object frame. The low-level policy is then trained to map from the canonicalized object point cloud and skill label to the canonicalized trajectory.
    During testing (right), given a task description and multi-view RGBD frames, the high-level agent generates a sequence of skills with associated object descriptions. For each step, it outputs an object mask and a skill label; the mask is lifted into a canonicalized object point cloud via the mask-as-frame interface, and together with the skill label, is provided to the low-level policy, which predicts the corresponding trajectory in the object frame for execution.
    } 
    \label{fig:teaser}
    \vspace{-0.7cm}
    \end{center}
\end{figure*}
Our approach is a two-level policy that learns object-centric skills and composes them to solve novel tasks. The high-level agent uses a VLM together with vision foundation models (VFM) to propose a step-wise sequence of (skill, object) pairs, and then the low-level controller predicts a trajectory in the object frame and maps it to the world frame for execution.
   The overview of our framework is shown in \autoref{fig:teaser}. Concretely, we first decompose task demonstrations into object-centric skills using the high-level agent, then train a diffusion policy controller to produce object-frame trajectories from an object point cloud and a skill token (Section~\ref{sec:ocdiscovery}). At test time, our skill inference lifts the predicted mask (from the high level) to a point cloud, applies the same canonicalization, queries the low-level module for the object-frame trajectory, and un-canonicalizes it for execution (Section~\ref{sec:skcomp}). This design allows high-level generalization to propagate to the low-level module, improving robustness to object geometry and spatial variations while keeping the interface simple and inspectable.

\subsection{Object-Centric Skill Discovery and Learning }
\label{sec:ocdiscovery}
Our goal at training time is to convert raw demonstrations into a library of reusable object-centric skills and to train a low-level controller that predicts skill trajectories in the object frame. Concretely, a skill is defined as $
(k, m, \tau_s)$, where $k \in \mathcal{K}$ is a skill label from the predefined skill library, 
$m = \{ m_v \in \{0,1\}^{H \times W} \mid v = 1, \dots, V \}$ is the multi-view binary mask across $V$ camera views to locate the target object, 
and $\tau_s = \{ s_1, \dots, s_n \} \in S^n$ is the interaction-relevant segment of the gripper trajectory (from near-approach through contact), excluding free-space approach to maximize transferability across scenes. 

\paragraph{High-level skill discovery} Given a task description $d$, a predefined skill label set $\mathcal{K}$, and multi-view RGB-D frames of a demonstration data, we first downsample the demonstration into keyframes (following~\cite{shridhar2023perceiver}) and further segment it into consecutive intervals using VLM, labeled as either free-space approach or a skill label $k\in\mathcal{K}$ associated with an object's linguistic description (e.g., ``red button on the table''). We then obtain the object mask $m$ by querying detection and segmentation models (DINOv2~\cite{oquab2023dinov2} followed by SAM2~\cite{ravi2024sam}). Each segment becomes a skill triplet $(k, m, \tau_s)$. The high level automatically flags low-confidence segments for optional human verification to ensure a clean training set.

\paragraph{Low-level training using mask-as-frame}

For each skill $(k,m,\tau)$, we lift $m$ into a 3D object point cloud $P=\{p_j\in\mathbb{R}^3\}_{j=1}^n$ using depth and calibrated extrinsics. We then \emph{canonicalize} both the point cloud and the action (in the world frame) into the object frame. Specifically, a randomly\footnote{This random sample of anchor point serves as a form of data augmentation.} sampled on-object anchor $t$ in the world frame defines the local frame center. We recenter $P$ and $\tau$ to obtain the canonicalized pair $(P_c,\tau_c)$, where $P_c = P - t; \tau_c = \tau - t$ are now in the object frame. We call this strategy of using an object mask to identify an object-centric coordinate frame \emph{mask-as-frame} $C_m$, and $C_m^{-1}$ its inverse, where $C_m(P) = P_c; C_m^{-1}(\tau_c) = \tau$. 

We thus obtain an \textbf{object-centric skill} as the triplet $(k, P_c, \tau_c)$ (with $(k, P_c)$ used during testing when $\tau_c$ is to be inferred). The skill label $k$ is encoded with a pretrained language model into $f_k\in\mathbb{R}^d$. We instantiate the low-level policy with an equivariant diffusion controller (EquiDiff~\cite{wang2024equivariant}) equipped with an observation encoder that consumes $(P_c,f_k)$ and predicts object-centric trajectory $\tau_c$ with a standard noise-prediction objective. Through the above formulations, the controller learns a library of canonical skill primitives that are agnostic to global object placement yet precise in the object's local geometry during training. To make the train and test contract explicit, we train
\(
\tau_c \;\leftarrow\; \pi_{\text{low}}\!\big(C_m(P),\, f_k\big),
\)
so that skills are learned in the object frame and can later be rebound to new objects via the same operator.

\subsection{Skill Composition at Inference}
\label{sec:skcomp}
At test time, we compose the learned object-centric skills by repeatedly binding a skill label to a target object and executing the resulting object-frame trajectory. Specifically, given the current observation, the high-level agent $\pi_{\text{high}}$ proposes the next skill $k$ (and its embedding $f_k$) and object mask $m$. Similar to training, we lift the object mask to a 3D object point cloud canonicalized into the object frame, $P_c = C_m(P)$, using the same object-frame transform $C_m$ as in training. 

The pair $(P_c,f_k)$ is fed to $\pi_{\text{low}}$ to predict the canonical trajectory $\tau_c$. We then map back to the world frame with the inverse transform,
\[
\hat{\tau} \;=\; C_m^{-1}\!\Big(\pi_{\text{low}}\!\big(C_m(P),\, f_k\big)\Big),
\]
and execute $\hat{\tau}$ on the robot. A motion planner connects the end of the current skill and the start of the next skill, decoupling the free-space approach from object-related execution.

\subsection{Design Choices}
In this section, we illustrate how our skill boundary, canonicalized skill representation, and using object-centric skills as the interface between the high- and low-level agents improve system-level generalization.

\paragraph{Skill Boundary}

We define a skill as the portion of the trajectory spanning from close-range object interaction, similar to demonstration generation pipelines~\cite{mandlekar2023mimicgen,xue2025demogen}, though our definition is intended for skill learning and composition. This boundary isolates core manipulation behavior from broader navigation, which in our framework is delegated to a motion planner, thereby significantly reducing learning complexity. It also enhances robustness by minimizing sensitivity to the robot’s initial configuration. For example, when skills are executed in a different sequence, the robot’s starting state is often out-of-distribution (OOD), which can lead to execution failures. Moreover, such segments can be automatically extracted using existing vision-language models (VLMs), enabling scalable and efficient skill representation.
In contrast, prior work such as ~\cite{garcia2025towards} defines a skill as the full trajectory from the gripper’s initial pose to task completion, which entangles global navigation with local manipulation and increases policy complexity.

\paragraph{Canonicalization}
After defining skills from the close-range object interaction to post-contact, we further apply canonicalization by aligning the object and its corresponding trajectory to the object frame. This frame transfer unifies the distribution of skill demonstrations, reduces data variance, and simplifies the learning problem for the low-level controller. More importantly, this formulation introduces object-level symmetry, allowing learned skills to generalize across different object poses, instances, and configurations with minimal retraining.

\paragraph{Interface Between High-Level and Low-Level Modules}
We use object-centric skills (represented as a skill label and the canonicalized object mask) as the interface. The object mask serves as a spatial reference frame, providing strong but flexible locality and enabling the low-level policy to operate in a consistent, object-aligned coordinate system. This relieves the low-level module from reasoning over the entire scene and improves robustness and modularity during skill execution.
Previous methods have explored alternatives such as image-space heatmaps~\cite{li2025bridgevla}, semantic masks~\cite{garcia2025towards}, or direct action outputs~\cite{xian2023chaineddiffuser}, each facing trade-offs in precision, interpretability, or stability.
\paragraph{Generalization Analysis}
An important advantage of the above design choices is that they support generalization across three key dimensions. First, \emph{spatial generalization} is achieved by enforcing object-level symmetry: the low-level controller receives only the point cloud of the manipulated object in its local frame, ensuring equivariance at the object level and allowing skill execution to remain consistent under spatial transformations of the target object. Second, \emph{instance and appearance generalization} arises from the fact that the low-level policy operates on object-centric observations derived from the mask and lifted into a point cloud. This encourages the controller to focus on local geometry rather than global appearance, enabling transfer to novel object instances within the same category and robustness to variations in color, texture, or background. Finally, \emph{skill compositional generalization} is supported by representing each skill relative to the object frame, allowing the high-level agent to reliably compose and reuse learned skills in novel long-horizon tasks. 

\section{Experiments}

\begin{table*}[ht]
\caption{\textbf{Performance comparison across training tasks.} Reported numbers are task success rates (\%). The \textbf{Mean} column averages all task scores for each method. Bold indicates the best score per column. \emph{Training Demo} indicates the number of demonstrations used.}
\label{tab:training}
\centering
\scalebox{0.89}{
\begin{tabular}{@{}lcccccccccccc@{}}
\toprule
\textbf{Method} & \textbf{Training Demo} & \textbf{Mean} & 
\makecell{Close\\Jar+15} & 
\makecell{Close\\Jar+16} & 
\makecell{LightBulb\\In+17} & 
\makecell{LightBulb\\In+19} & 
\makecell{Pick\\Lift+0} & 
\makecell{Pick\\Lift+2} & 
\makecell{Pick\\Lift+7} & 
\makecell{Push\\Btn+0} & 
\makecell{Push\\Btn+3} & 
\makecell{Open\\Drawer+0} \\ 
\midrule
Hiveformer      & 100 & 69.1 & 64  & 92  & 12  & 13  & 86  & 92  & 93  & 84  & 68  & 15  \\
PolarNet          & 100 & 90.4 & 99  & 99  & 72  & 71  & 92  & 84  & 88  & \textbf{100} & \textbf{100} & 61  \\
3D Diffuser Actor & 100 & 96.2 & \textbf{100} & \textbf{100} & 85  & 88  & 99  & 99  & 99  & 98  & 96  & 82  \\
RVT-2                 & 100 & 97.6 & 97  & 98  & \textbf{93}  & 91  & 99  & 98  & \textbf{100} & \textbf{100} & \textbf{100} & \textbf{93} \\
3D-LOTUS           & 100 & 96.5 & \textbf{100} & \textbf{100} & 84  & 85  & 99  & \textbf{100} & 99  & 99  & 99  & 83 \\
3D-LOTUS++       & 100 & 88.3 & \textbf{100} & 99  & 55  & 45  & 97  & 94  & 93  & \textbf{100} & \textbf{100} & 68 \\
BridgeVLA        & 100 & 96.8 & 98  & \textbf{100} & 90  & 87  & 99  & \textbf{100} & 98  & \textbf{100} & 98  & 86  \\
Ours                              & \textbf{3} & \textbf{98.0} & \textbf{100} & \textbf{100} & 90  & \textbf{95}  & \textbf{100} & \textbf{100} & \textbf{100} & 95  & \textbf{100} & 90 \\
\bottomrule
\end{tabular}
}
\end{table*}

\begin{table*}[ht]
\caption{\textbf{Zero-shot transfer performance on testing tasks.} Reported values are success rates (\%). \emph{Training Demo} indicates the number of demonstrations used in the training task. Boldface denotes the best score in each column.}
\label{tab:testing}
\centering
\scalebox{0.90}{
\begin{tabular}{@{}lcccccccccccccc@{}}
\toprule
\textbf{Method} & \textbf{Training Demo} & \textbf{Mean} &
\makecell{Push\\Btn+13} & \makecell{Push\\Btn+15} & \makecell{Push\\Btn+17} &
\makecell{Pick\\Lift+14} & \makecell{Pick\\Lift+16} & \makecell{Pick\\Lift+18} &
\makecell{Pick\\Lift\\Star+0} & \makecell{Pick\\Lift\\Moon+0} & \makecell{Pick\\Lift\\Toy+0} &
\makecell{Close\\Jar+3} \\  
\midrule
Hiveformer        & 100 & 37.5& 97 & 85 & 88 & 21 & 9 & 8 & 73 & 88 & 87 & 0 \\
PolarNet            & 100 & 46.5 & \textbf{100} & \textbf{100} & 85 & 3 & 1 & 0 & 88 & 93 & 90 & 20 \\
3D Diffuser Actor & 100 & 32.4 & 87 & 81 & 60 & 9 & 18 & 0 & 43 & 91 & 30 & 23 \\
RVT-2                   & 100 & 46.5 & \textbf{100} & \textbf{100} & \textbf{100} & 47 & 29 & 8 & \textbf{98} & 94 & 78 & 7 \\
3D-LOTUS             & 100 & 46.7 & 99 & \textbf{100} & \textbf{100} & 3 & 18 & 33 & 69 & 80 & \textbf{96} & 71 \\
3D-LOTUS++         & 100 & 66.8 & 99 & \textbf{100} & 99 & 94 & \textbf{96} & \textbf{95} & 94 & 29 & 71 & \textbf{98} \\
BridgeVLA          & 100 & 58.9 & \textbf{100} & \textbf{100} & \textbf{100} & 74 & 89 & 0 & 99 & 95 & 93 & 66 \\
Ours                                & \textbf{3}   & \textbf{82.3} & \textbf{100} & \textbf{100} & \textbf{100} & \textbf{100} & \textbf{100} & 95 & \textbf{100} & \textbf{100} & 95 & 90 \\
\bottomrule
\end{tabular}
}

\vspace{1ex}

\scalebox{0.87}{
\begin{tabular}{@{}lcccccccccccc@{}}
\toprule
\textbf{Method} &
\makecell{Close\\Jar+4} &
\makecell{LightBulb\\In+1} & \makecell{LightBulb\\In+2} &
\makecell{Lamp\\On+0} & \makecell{Push\\Btns4+1} &
\makecell{Push\\Btns4+2} & \makecell{Push\\Btns4+3} &
\makecell{Open\\drawer2+0} & \makecell{Open\\drawer3+0} & \makecell{Open\\drawer+1} & \makecell{Open\\drawer\\long+0} & \makecell{Open\\drawer\\long+1} \\ 
\midrule
Hiveformer        & 0 & 4 & 0 & 7 & 0 & 0 & 0 & {59} & {39} & {0} & {78} & {82} \\
PolarNet            & 82 & 22 & 17 & 14 & 1 & 0 & 0 & \textbf{91} & {39} & {4} & \textbf{84} & \textbf{88} \\
3D Diffuser Actor & 82 & 51 & 60 & 7 & 0 & 0 & 0 & {19} & {1} & {0} & {15} & {35} \\
RVT-2                   & 77 & 68 & 6 & 0 & 0 & 0 & 0 & {81} & {0} & \textbf{6} & \textbf{84} & {39} \\
3D-LOTUS             & 90 & 24 & 41 & 0 & 3 & 0 & 0 & {90} & {22} & {0} & {56} & {33} \\
3D-LOTUS++         & 96 & 56 & 43 & 2 & 76 & 49 & 37 & {70} & {41} & {0} & {72} & {52} \\
BridgeVLA          & 88 & 66 & 74 & 7 & 0 & 0 & 0 & {65} & \textbf{87} & {0} & {59} & {34} \\
Ours                                & \textbf{100} & \textbf{90} & \textbf{85} & \textbf{50} & \textbf{95} & \textbf{90} & \textbf{85} & {90} & {85} & {0} & {60} & {0} \\
\bottomrule
\end{tabular}
}
\end{table*}

To evaluate our framework, we conduct experiments in both simulation and real-world environments.
\subsection{Simulation Experiment}
\paragraph{Experimental Setup}
\begin{figure}[t]
    \centering
    \includegraphics[width=\columnwidth]{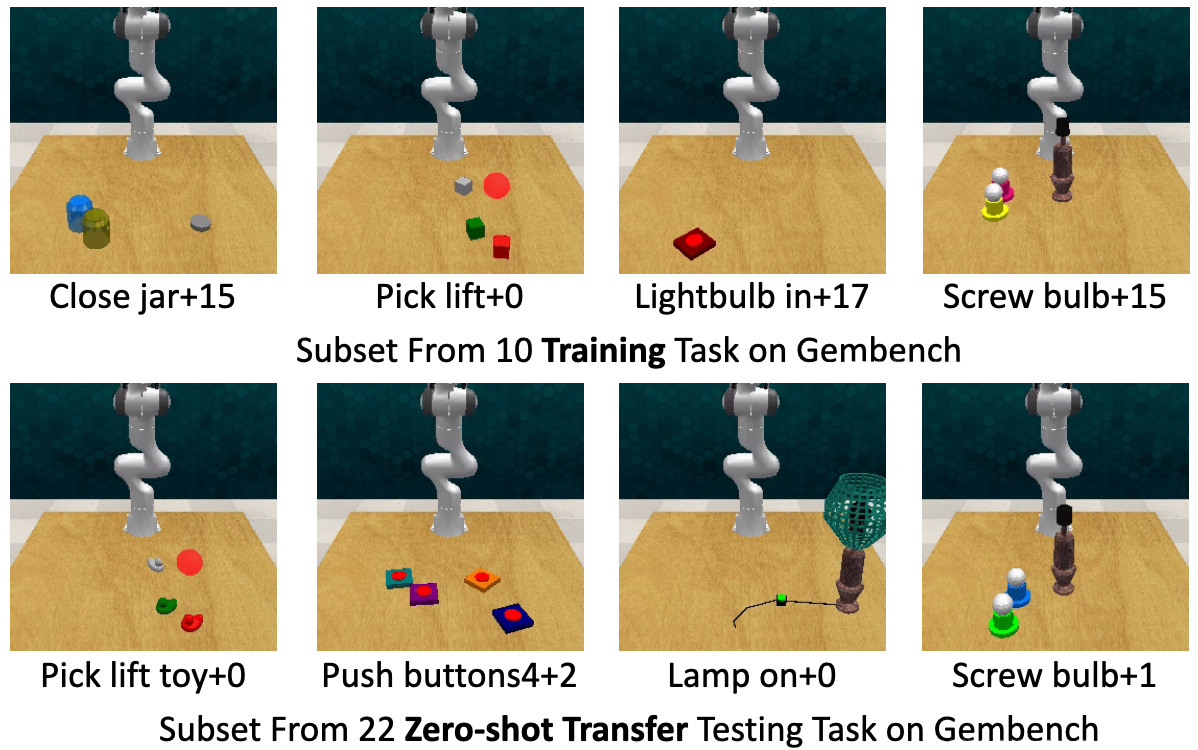}
    \caption{\textbf{Subset of the full 10 training tasks and 22 zero-shot transfer testing tasks.} Zero-shot transfer testing tasks include unseen appearances (novel instances and colors) as well as unseen task types. In each task, the initial scene configuration is randomized, introducing spatial variations across both demonstrations and testing settings. }
    \label{fig:sim}
    \vspace{-0.8cm}
\end{figure}
We evaluate our policy in simulated environments using the GemBench benchmark~\cite{garcia2025towards}, implemented with CoppeliaSim~\cite{coppeliaSim} and PyRep~\cite{james2019pyrep}. 
The simulation setup consists of a 7-DoF Franka Panda robot with a parallel gripper, observed by two RGB-D cameras.  We train GSL on 10 GemBench training tasks involving \textit{press}, \textit{pick}, \textit{screw}, and \textit{lift} skills. 
For evaluation, we test on both the 10 training tasks but with spatial variation and 22 testing tasks with spatial, object appearance, and compositional variation to probe generalization. 
A subset of these training and testing tasks is visualized in \autoref{fig:sim}. 
For example, the testing task \textit{Lamp On+0} requires transferring the pressing skill learned from \textit{Push
Btn+0 and +3} during training. To further challenge generalization, we train GSL using only 3 demonstrations per training task, compared to the standard 100 demonstrations used by baselines. 
Each task is evaluated over 20 rollouts, and the average success rate is reported. 

\paragraph{Baselines}
We compare GSL against seven state-of-the-art methods on GemBench: 
two 2D image-based policies (HiveFormer~\cite{hiveformer} and RVT-2~\cite{rvt2}), 
three 3D-based models (PolarNet~\cite{polarnet}, 3D Diffuser Actor~\cite{ke20243d}, and 3D-Lotus~\cite{garcia2025towards}), 
and two hierarchical methods that incorporate VLM/LLMs and/or pretrained vision foundation models (BridgeVLA~\cite{li2025bridgevla} and 3D-Lotus++~\cite{garcia2025towards}). We'll use the result in original paper if possible.

\paragraph{Results}
As shown in \autoref{tab:training} and \autoref{tab:testing}, GSL achieves a 1.2\% improvement on training tasks and a substantial 23.4\% gain on testing tasks, despite using only 3 demonstrations compared to 100 in the baseline. Specifically, GSL outperforms baselines by 1.2\% on tasks requiring only spatial generalization, by 12.2\% on tasks requiring both instance and appearance generalization and spatial generalization, and by 36.0\% on tasks requiring full spectrum of instance/appearance, spatial, and skill compositional generalization. 
The main failure case occurs in drawer-related tasks, where the high-level agent segments the entire drawer cabinet as a single object, rather than identifying individual drawers. This forces the low-level policy to infer which drawer to interact with, violating our design principle of keeping semantic reasoning within the high-level agent. We attribute this limitation to the high-level visual foundation model (VFM), particularly SAM2, which lacks part-level segmentation capability. Incorporating part-aware segmentation or fine-tuning the high-level agent could help mitigate this issue. 


\newcolumntype{l}{>{\raggedright\arraybackslash}p{2cm}}
\newcolumntype{c}{>{\centering\arraybackslash}p{1.2cm}}
\begin{table*}[!t]
\caption{\textbf{Performance comparison across real-world tasks.}}
\label{tab:realworld}
\renewcommand\arraystretch{0.8}
    \setlength{\tabcolsep}{11pt}
    \centering
    \vspace{-0.1cm}
    \begin{tabular}{l c c c c c c c }
        \toprule
         Method &Demo& Toy in cup & Toy in box & Hammer from Drawer & Teabag in cup & Teabag in red cup & Prepare Tea\\
        \midrule
        3D-LOUTS++ & 30 & 7/10 & 4/10 & 5/10 & 6/10 & 4/10&3/10 \\
        Ours &\textbf{3}& \textbf{9/10} & \textbf{9/10} & \textbf{6/10} & \textbf{9/10} & \textbf{9/10} & \textbf{6/10} \\
        \hline
        
    \end{tabular}
    \label{tab:specialcase_real}
    \vspace{-0.5cm}
\end{table*}
\subsection{Ablation Study}

\begin{table}[ht]
\caption{\textbf{Performance of Different Ablations on Various Tasks.}}
\label{tab:ablation-results-all}
\vskip 0.15in
\centering
\setlength{\tabcolsep}{0.3pt}
\scriptsize
\newcolumntype{C}{>{\centering\arraybackslash}p{0.9cm}} 
\begin{tabular}{@{}lCCCCCCC@{}}
\toprule
Method & Mean & Push Btn+0 & LightBulb
In+17 & Push Btns4+3 & Lamp On+0 & LightBulb
In+1 \\
\midrule
Regular Skill & 0.37 & 0.65 & 0.5 & 0.2 & 0.1 & 0.4   \\
No Canonic.              & 0.12 & 0.3 & 0.1 & 0 & 0.1 & 0.1  \\
Action as Interf.              & 0.1 & 0.25 & 0.1 & 0 & 0.05 & 0.1  \\
Heatmap as Interf.              & 0.17 & 0.5 & 0.05 & 0.1 & 0.2 & 0  \\
\textbf{Complete Model}        & \textbf{0.83} & \textbf{1} & \textbf{0.9} & \textbf{0.85} & \textbf{0.50} & \textbf{0.9}  \\
\bottomrule
\end{tabular}
\vspace{-0.5cm}
\end{table}
To evaluate the contributions of each design component, we conduct an ablation study across two training tasks and three testing tasks. The test set covers: (1) tasks requiring only spatial generalization (\textit{Push Btn+0} and \textit{LightBulb In+17}), (2) tasks requiring both instance and appearance generalization and spatial generalization (\textit{Lamp On+0} and \textit{LightBulb In+1}), and (3) tasks requiring  skill compositional generalization, along with appearance and spatial generalization (\textit{Push Btns4+3}). We compare the following configurations: \textbf{Regular Skill}, which includes the gripper approach phase as part of the skill and uses the full robot state as input to the low-level controller; \textbf{No Canonic.}, which disables canonicalization; \textbf{Action as Interf.}, which uses robot keyposes as the interface between high- and low-level modules; and \textbf{Heatmap as Interf.}, which uses heatmaps appended to visual inputs to indicate the target object. As shown in \autoref{tab:ablation-results-all}, removing or altering any key design choice leads to a consistent performance drop. Our object-centric skill definition (excluding the approach phase) improves performance across all tasks, most significantly under skill compositional generalization (\textit{Push Btns4+3}), where it provides a 65\% boost. Similarly, removing canonicalization or replacing the interface degrades performance, highlighting the importance of both components for robust generalization.


\subsection{Real World Experiments}
\begin{figure}[h!]
    \centering
    \includegraphics[width=\columnwidth]{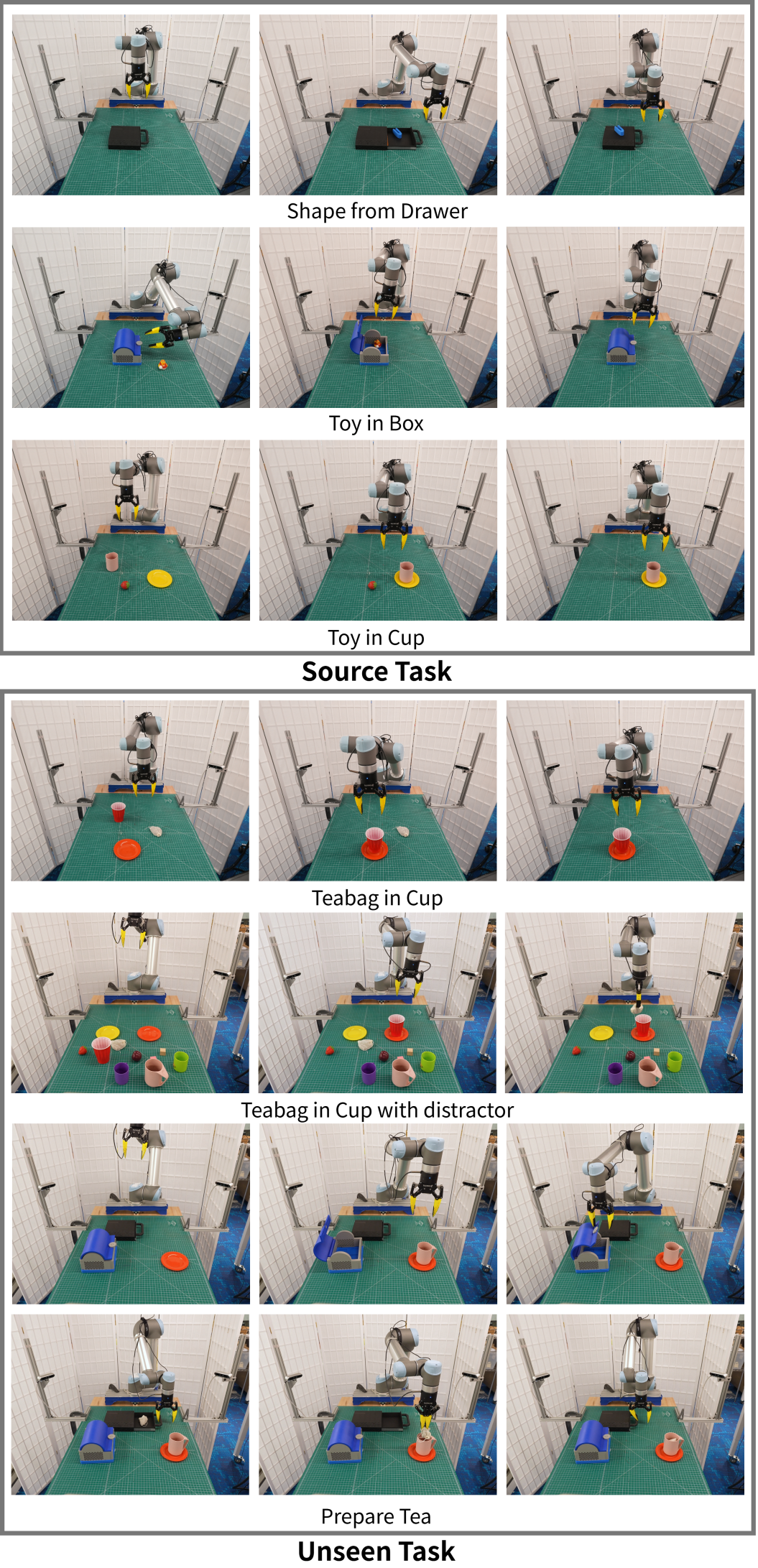}
    \caption{\textbf{Six real-world tasks, including three source tasks and three unseen tasks.}}
    \label{fig:realworld}
    \vspace{-0.5cm}
\end{figure}

We further evaluate the performance and generalization of GSL on a physical system consisting of a UR5 robot with a parallel gripper and three Intel RealSense~\cite{Keselman_2017_CVPR_Workshops} D455 RGBD cameras. 
All policies are trained on three source tasks: \textit{Shape from Drawer}, where the robot pull out a drawer, retrieves a U-shaped block, places it on the table, and push back the drawer; \textit{Toy in Box}, where the robot opens a box, places a toy duck inside, and closes the lid; and \textit{Toy in Cup}, where the robot first places a pink cup onto an orange plate and then places a toy strawberry into the cup. 

To evaluate generalization, the trained policies are applied directly, without any additional finetuning, to a suite of six testing tasks that extend the training distribution along several dimensions, including spatial variation, object appearance changes, environmental distractors, and novel long-horizon compositions. The evaluation begins with three spatially modified versions of the source tasks, testing robustness to positional changes. It then introduces \textit{Teabag in Cup}, in which the robot places a white teabag into a red cup situated on a yellow plate, requiring transfer of the \textit{pick} skill to a novel object instance and the \textit{place} skill to a plate with an unseen color. \textit{Teabag in Cup with Distractor} increases complexity by introducing multiple distractor cups and toys, including a pink cup and a toy strawberry identical to the training instances used for the \textit{place} and \textit{pick} skills, thereby evaluating robustness against distractor interference. Finally, \textit{Prepare Tea} involves executing a long sequence of previously learned skills, \textit{pick}, \textit{place}, \textit{push back}, \textit{pull out}, \textit{open}, and \textit{close}, in a coordinated manner to simulate the process of preparing tea. This task integrates interactions with the drawer, box, cups, and teabag into a coherent routine. 

All tasks are illustrated in \autoref{fig:realworld}. As a stretch goal for data efficiency, our policy is trained on only three demonstrations per source task, whereas the \textit{3D-Lotus++} baseline uses 30 demonstrations.

\paragraph{Results}
As shown in \autoref{tab:realworld}, GSL trained with just 3 demonstrations outperforms baseline methods trained with 30. It achieves a 26.7\% improvement on tasks requiring spatial generalization, a 40.0\% gain on tasks involving both instance/appearance and spatial generalization, and a 30.0\% improvement on tasks demanding the full spectrum of instance/appearance, spatial, and skill compositional generalization. Most failure cases are attributed to inconsistent high-level segmentation, which was not fine-tuned. For example, when executing the "pull-out" skill on a drawer, the high-level agent often failed to identify the handle as part of the drawer, likely due to missing fine-grained part annotations in the VFM (SAM) pretraining data. When ground-truth masks were manually provided, success rates increased by 20\% on \textit{Prepare tea} and 30\% on \textit{Hammer from Drawer}, highlighting high-level segmentation as a promising direction for future improvement.


\section{Conclusion and Limitation}
\label{sec:limit}
In this paper, we propose GSL, a novel hierarchical policy learning framework. By utilizing object-centric skills as an interface, our policy achieves  significantly higher generalizability and sample-efficiency compared with the baselines in both simulation and the real world. One major limitation of our work is that
our formulation focuses on object-centric skills involving only a single manipulated object. Although this scope covers a wide range of manipulation primitives (e.g., opening a drawer), it does not explicitly handle multi-object interactions, such as pushing a block to a designated base area, where reasoning about multiple objects is required. Future work can extend GSL to more general multi-object interactions by incorporating multiple masks into the mask-as-frame interface. Another limitation is that our approach assumes the high-level can reliably provide accurate object masks. While this assumption is reasonable given the robustness of vision-language models trained on large-scale data, imperfect grounding or segmentation may reduce performance. Addressing robustness to noisy or incorrect masks is left for future study.

\vspace{-3mm}

\bibliographystyle{IEEEtran}
\bibliography{compact_bib}
\end{document}